\documentclass[sigconf]{acmart}
 \usepackage{subfigure}
 \newcommand{\redfont}[1]{{\color{red}{#1}}}
 \newcommand{\bluefont}[1]{{\color{blue}{#1}}}
 
 \usepackage{multirow}
 \usepackage{hyperref}

\usepackage{balance}

\def\BibTeX{{\rm B\kern-.05em{\sc i\kern-.025em b}\kern-.08emT\kern-.1667em\lower.7ex\hbox{E}\kern-.125emX}}
    
\copyrightyear{2019}
\acmYear{2019}
\setcopyright{acmcopyright}
\acmConference[MM '19] {Proceedings of the 27th ACM International Conference on Multimedia}{October 21--25, 2019}{Nice, France}
\acmBooktitle{Proceedings of the 27th ACM International Conference on Multimedia (MM'19), Oct. 21--25, 2019, Nice, France}
\acmPrice{15.00}
\acmDOI{10.1145/3343031.3351074}
\acmISBN{978-1-4503-6889-6/19/10}

\begin{document}

\fancyhead{}

\title{Knowledge-guided Pairwise Reconstruction Network for Weakly Supervised Referring Expression Grounding}


\author{Xuejing Liu$^{1,2}$, Liang Li$^{1, *}$, Shuhui Wang$^{1}$, Zheng-Jun Zha$^{3}$, Li Su$^{2,1}$, Qingming Huang$^{2,1}$}
\thanks{$^\ast$Corresponding author.}
\affiliation{%
	\institution{$^1$Key Laboratory of Intelligent Information Processing, Institute of Computing Technology, CAS, Beijing, China}
	\institution{$^2$School of Computer Science and Technology, University of Chinese Academy of Sciences, Beijing, China}
	\institution{$^3$School of Information Science and Technology, University of Science and Technology of China, Hefei, China}
}
\email{{xuejing.liu, liang.li}@vipl.ict.ac.cn, wangshuhui@ict.ac.cn, zhazj@ustc.edu.cn, {suli, qmhuang}@ucas.ac.cn}

%
\renewcommand{\shortauthors}{Trovato and Tobin, et al.}

%
\begin{abstract}
Weakly supervised referring expression grounding (REG) aims at localizing the referential entity in an image according to linguistic query, where the mapping between the image region (proposal) and the query is unknown in the training stage. 
In referring expressions, people usually describe a target entity in terms of its relationship with other contextual entities as well as visual attributes. However, previous weakly supervised REG methods rarely pay attention to the relationship between the entities. In this paper, we propose a knowledge-guided pairwise reconstruction network (KPRN), which models the relationship between the target entity (subject) and contextual entity (object) as well as grounds these two entities.
Specifically, we first design a knowledge extraction module to guide the proposal selection of subject and object. 
The prior knowledge is obtained in a specific form of semantic similarities between each proposal and the subject/object.
Second, guided by such knowledge, we design the subject and object attention module to construct the subject-object proposal pairs.
The subject attention excludes the unrelated proposals from the candidate proposals. The object attention selects the most suitable proposal as the contextual proposal. 
Third, we introduce a pairwise attention and an adaptive weighting scheme to learn the correspondence between these proposal pairs and the query.  Finally, a pairwise reconstruction module is used to measure the grounding for weakly supervised learning.
Extensive experiments on four large-scale datasets show our method outperforms existing state-of-the-art methods by a large margin\footnote{Code is available at {\url{https://github.com/GingL/KPRN}}.}.
\end{abstract}

%
%


%

%
\maketitle

\begin{figure}
	\centering
	\includegraphics[width=8cm]{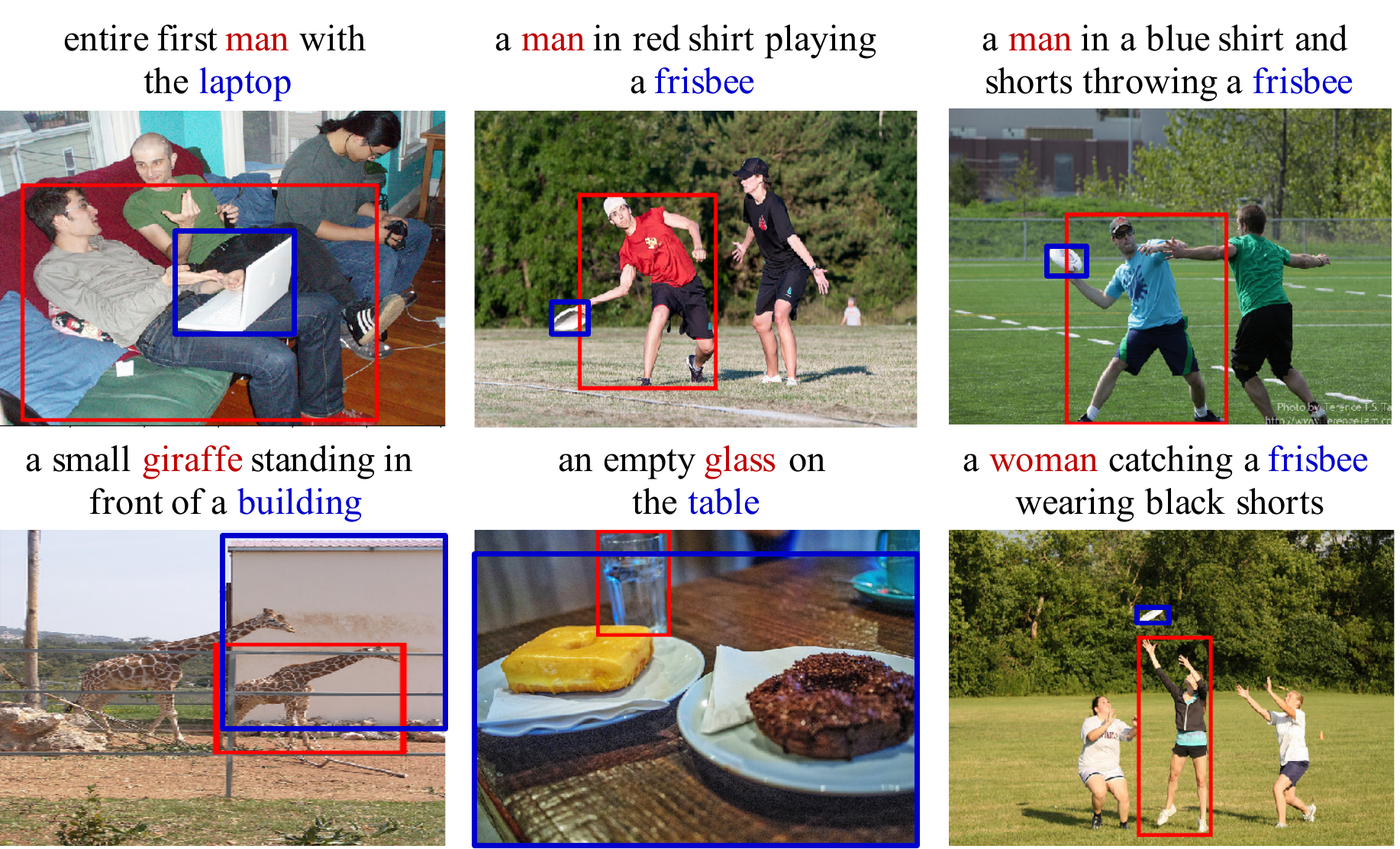}
	\caption{Examples of Referring Expression Grounding (REG). Given a query and an image, REG aims to localize the referential entity. We can observe that besides visual attributes, relationship with other entities is often used to describe the target entity. 
		The target entity and its corresponding proposal are shown in \redfont{red}. The contextual entity and its corresponding proposal are shown in \bluefont{blue}.
	}
	\label{examples}
\end{figure}

\section{Introduction}
Referring expression grounding (REG), also known as phrase localization, has been a surge of interest in both computer vision and natural language processing \cite{DBLP:conf/cvpr/MaoHTCY016, DBLP:conf/cvpr/HuXRFSD16, DBLP:conf/eccv/RohrbachRHDS16, DBLP:conf/cvpr/Yu0SYLBB18, 8695120,DBLP:conf/mm/0003WJH18}. Given a query (referring expression) and an image, REG is to find the corresponding location in the image of the referential entity.  
REG can be widely used in interactive applications, such as robotic navigation \cite{DBLP:conf/acl/ThomasonSM17, DBLP:conf/cvpr/AndersonWTB0S0G18, DBLP:conf/cvpr/WangJHT12}, visual Q\&A \cite{DBLP:conf/cvpr/DasDGLPB18, DBLP:conf/cvpr/GordonKRRFF18, DBLP:journals/ieeemm/LiJZWH13}, or photo editing \cite{DBLP:journals/tog/ChengZLVSCMT14, DBLP:conf/mm/WangCZH018}.
Traditionally, training the REG model in a supervised manner requires expensive annotated data that explicitly draw the connection between the input query and its corresponding proposal in the image. Besides, supervised REG models can only handle the grounding with certain categories in the limited training data, which cannot meet the demand for real-world applications. Hence we focus on weakly supervised REG task, where only the image-query pairs are used for training without the mapping information between the query and the proposal.

The weakly supervised REG problem can be formulated as follows. Given an image $I$, a query $q$ and a set of region proposals $\{r_i\}^N_{i=1}$, REG aims at selecting the best-matched region $r^*$ according to the query without the ground-truth pair $(q, r_i)$ . To find the correct mapping between the query and proposal under weakly supervised scenario, 
Rohrbach {\it et al.}~\cite{DBLP:conf/eccv/RohrbachRHDS16} select the best proposal from a set of candidate proposals through attention mechanism, then reconstruct the input query based on the selected proposal.
Chen {\it et al.}~\cite{DBLP:conf/cvpr/ChenGN18} design knowledge aided consistency network, which reconstructs both the input query and proposal's information. 
Xiao {\it et al.}~\cite{DBLP:conf/cvpr/XiaoSL17} generate attention mask to localize linguistic query based on image-phrase pairs and language structure.
Zhao {\it et al.}~\cite{DBLP:conf/cvpr/0006LZF18} try to find the location of the referential object by searching over the entire image. 
All the above methods only exploit the visual attribute features of proposals during grounding and reconstruction. 

However, in addition to the visual attribute information, people often describe a target entity in terms of its relationship with other contextual entities, as shown in Fig. \ref{examples}. 
Recently, Zhang \textit{et al.} \cite{DBLP:conf/cvpr/ZhangNC18} propose a variational Bayesian method to exploit the reciprocal relation between the referent and context for referring expression grounding. 
However, their method learns to model the relationship between referent and context based on the annotation of the target entity. Thus the model is not suitable under the weakly supervised setting, where neither the annotations for target nor that for context are available for training. 

To address the above challenge, we propose a knowledge-guided pairwise reconstruction network (KPRN) for weakly supervised REG. KPRN learns the mapping between proposal pair (subject, object) and query with the assistance of prior knowledge, and grounds these two entities.  KPRN mainly consists of the following three steps. 

First, to fulfill the lack of annotations for target entity (subject) and contextual entity (object), we design a knowledge extraction module to guide the proposal selection of subject and object.
Specifically, subject and object are first extracted from the input query. The category for each candidate proposal is obtained through Faster R-CNN \cite{DBLP:conf/nips/RenHGS15}. Then, both the proposal category and the subject/object are encoded into embeddings. 
Subsequently, the prior knowledge is obtained in a specific form of semantic similarities between the proposal and the subject/object.

Second, under the guidance of such knowledge, we design the subject and object attention module to construct the subject-object proposal pairs.
Subject attention learns to discard the unrelated candidate subject proposals. Object attention learns to select the best proposal as the contextual proposal. 

Third, we introduce pairwise attention to learn the matching score, which represents the correspondence between these proposal pairs and the query. Further, we design an adaptive weighting scheme for refining the correspondence based on the spatial relationship in the subject-object proposal pair. 
As the measurement of weakly supervised grounding, a pairwise reconstruction module is used to reconstruct the input query based on the subject-object proposal pairs and their matching scores.

KPRN can be trained in an end-to-end manner. At the inference stage, KPRN only utilizes the grounding to localize the referent without reconstruction. 

The main contributions of this paper are concluded as follows: 
\begin{itemize}
	\item We propose an end-to-end knowledge-guided pairwise reconstruction network, which models the mapping between the input query and proposal pair (subject, object), and grounds the subject and object. We design a knowledge extraction module to introduce the supervision of prior knowledge.
	\item We design the subject and object attention to compose the subject-object proposal pairs under the guidance of prior knowledge. The subject attention can exclude the unrelated candidate subject proposals, and the object attention can help find the best contextual proposal.   
	\item Through pairwise attention and an adaptive weighting scheme, the matching scores are learned to represent the correspondence between the subject-object proposal pairs and the query.
	A pairwise reconstruction is used to reconstruct the input query with attentive pairwise proposals.
	\item Comparison and ablation experiments on the RefCLEF and three MS-COCO datasets show that the proposed KPRN achieves state-of-the-art results in the weakly supervised REG task.
\end{itemize}

\begin{figure*}[ht]
	\includegraphics[width=0.78\textheight]{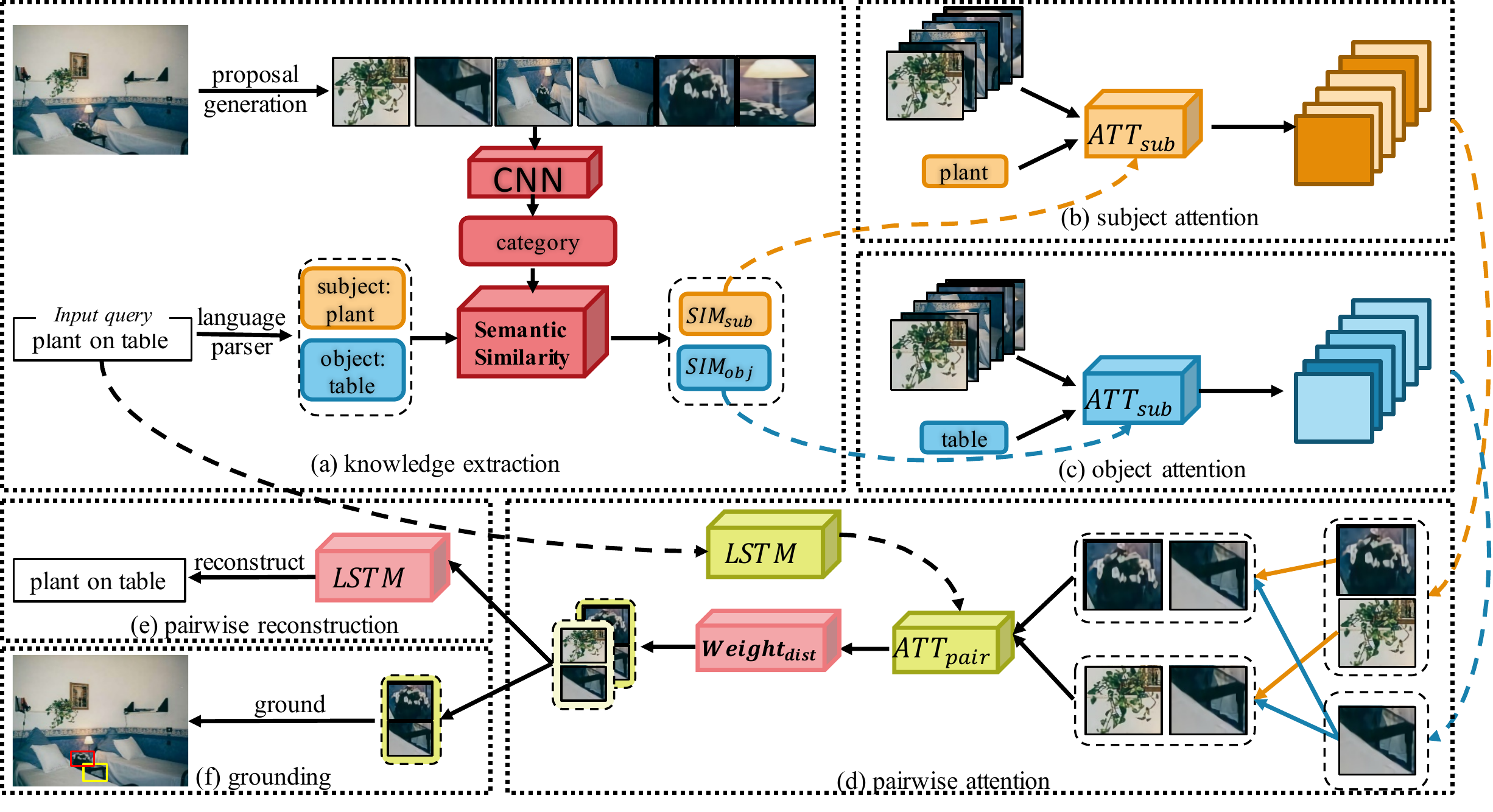}
	\caption{The framework of the proposed KPRN. It consists of (a) knowledge extraction (Section \ref{knowledge}), (b) subject attention (Section \ref{sub_att}), (c) object attention (Section \ref{sub_att}), (d) pairwise attention (Section \ref{pair_att}) (e) pairwise reconstruction (Section \ref{rec}) and (f) grounding (Section \ref{pair_att}). The composition of ($a\to b \to c\to d\to e$) is used for training, while the combination of ($a\to b \to c\to d\to f$) is used for inference. }
	\label{structure}
\end{figure*}
\section{Related Work}

\paragraph{Supervised Referring Expression Grounding (REG)}
REG \cite{DBLP:conf/emnlp/KazemzadehOMB14, DBLP:conf/naacl/MitchellDR13, DBLP:conf/emnlp/FitzGeraldAZ13, DBLP:conf/cvpr/MaoHTCY016, DBLP:conf/eccv/YuPYBB16,  DBLP:conf/cvpr/YuTBB17, DBLP:journals/corr/abs-1812-03426, DBLP:conf/cvpr/YangLWZH17} is also known as referring expression comprehension or phrase localization, which is the inverse task of referring expression generation.
REG intends to localize the corresponding object described by a free-form natural language query in an image.
Given an image $I$, a query $q$ and a set of region proposals $\{r_i\}^N_{i=1}$, REG selects the best-matched region $r^*$ according to the query.
Most REG methods can be roughly divided into two kinds.
One is CNN-LSTM based encoder-decoder structure to model $P(q|I,r)$ \cite{DBLP:conf/cvpr/MaoHTCY016, DBLP:conf/eccv/YuPYBB16, DBLP:conf/eccv/NagarajaMD16, DBLP:conf/cvpr/HuXRFSD16, DBLP:conf/cvpr/LuoS17, DBLP:journals/tip/SongWHT17}.
The other is the joint vision-language embedding framework to model $P(q, r)$.
During training, the supervision is object proposal and referring expression pairs $(r_i, q_i)$ \cite{DBLP:conf/eccv/RohrbachRHDS16, DBLP:conf/cvpr/WangLL16, DBLP:conf/iccv/Liu0017, DBLP:conf/iccv/ChenKN17, DBLP:conf/cvpr/Yu0SYLBB18, DBLP:journals/tmm/LiJH12}.
The relationship between the target entity and context entity is often used to assist grounding the target in supervised REG methods \cite{DBLP:conf/eccv/NagarajaMD16,DBLP:conf/iccv/PlummerMCHL17,DBLP:conf/cvpr/HuRADS17,DBLP:conf/cvpr/KrishnaCBF18,DBLP:conf/cvpr/ZhangNC18, DBLP:journals/ijon/LiYCZYJH16}. 
However, these methods learn to model the relationship between target entity and contextual entity based on the annotation of the target entity, which is not available under weakly supervised scenario.

	\paragraph{Weakly Supervised Referring Expression Grounding.}
	Weakly supervised REG only has image-level correspondence, and there is no mapping between image regions and referring expressions. To solve this problem,  Rohrbach {\it et al.}~\cite{DBLP:conf/eccv/RohrbachRHDS16} propose a framework which learns to ground by reconstructing the given referring expression through attention mechanism. Based on this framework, Chen {\it et al.}~\cite{DBLP:conf/cvpr/ChenGN18} design knowledge aided consistency network, which reconstructs both the input query and proposal's information. 
	Xiao {\it et al.}~\cite{DBLP:conf/cvpr/XiaoSL17} ground arbitrary linguistic phrase in the form of spatial attention mask and propose a network with discriminative and structural loss.
	Different from selecting the optimal region from a set of region proposals, Zhao {\it et al.}~\cite{DBLP:conf/cvpr/0006LZF18} propose multi-scale anchored transformer network, which can search the entire spatial feature map by taking region proposals as anchors to get a more accurate location. 
	

\section{Methodology}
Weakly supervised REG intends to ground the target entity described by the query under the scenario where the region-query correspondence is not available during training. 
To overcome the lack of supervised information, previous methods usually utilize the selected image regions to reconstruct the corresponding query. Here, we develop our method under such reconstruction mechanism.

Different from previous methods, we reconstruct the input query using subject-object proposal pairs, where subject proposal denotes the target proposal, and object proposal represents the context proposal.  Specifically, we propose a knowledge-guided pairwise reconstruction network (KPRN), which learns the mapping between proposal pairs and query with the assistance of prior knowledge, and grounds these two entities. The overall structure is shown in Fig. \ref{structure}.
Initially, through a knowledge extraction module, we introduce the supervision information for the selection of subject/object proposals. The supervision is prior knowledge obtained in a specific form of semantic similarities, representing the uniformity between the proposal category and the subject/object.
Secondly, with the supervision information, subject and object attentions are learned to exclude the unrelated proposals from the candidates and compose the proposal pairs.
Thirdly, pairwise attention and a weighting scheme are designed to learn the matching score to represent the correspondence between the query and the proposal pairs. Pairwise reconstruction is utilized to measure the grounding under weakly supervised scenario.

In the following, we first introduce the feature encoding of the image and query, and then detail every module of our method.

\subsection{Feature Encoding}
\label{feats_enc}

\subsubsection{Visual Features}
\label{visfeats}
Given an image with a set of region proposals, obtained by any off-the-shelf proposal generator \cite{DBLP:conf/eccv/ZitnickD14} or object detectors \cite{DBLP:conf/nips/RenHGS15}, we extract visual features for each proposal by the pre-trained networks and arithmetic operators. Here we use two kinds of visual features, including subject and object features.

\textbf{Subject feature} is the concatenation of CNN features and spatial representations for each proposal. Given an image $I$ and a set of region proposals $\{r_i\}^N_{i=1}$, we run the forward propagation of Faster R-CNN based on ResNet \cite{DBLP:conf/cvpr/HeZRS16} for each image $I$, and crop its C3 and C4 features as the CNN feature $\widetilde { v } _ { s } ^ { i } = f_{CNN}(r_i)$ for each proposal. The C3 features represent lower-level features such as colors and shapes while C4 features contain higher-level representations. 

Spatial representations consists of absolute position and relative locations with other entities of the same category in the image. Following \cite{DBLP:conf/eccv/YuPYBB16, DBLP:conf/cvpr/YuTBB17, DBLP:conf/cvpr/Yu0SYLBB18}, the absolute location feature of each proposal is decoded as a 5-dim vector $v_l^i = \left[ \frac { x _ { t l } } { W } , \frac { y _ { t l } } { H } , \frac { x _ { b r } } { W } , \frac { y _ { b r } } { H } , \frac { w \cdot h } { W \cdot H } \right]$, denoting the top-left, bottom-right position and relative area of the RoI to the whole image. The relative location feature indicates the relative location information between the target proposal and 5 surrounding proposals of the same category. 
For each surrounding proposal, we calculate its offset and area ratio to the candidate: $\delta v_l^{ij} = \left[ \frac { \left[ \Delta x _ { t l } \right] _ { i j } } { w _ { i } } , \frac { \left[ \triangle y _ { t l } \right] _ { i j } } { h _ { i } } , \frac { \left[ \triangle x _ { b r } \right] _ { i j } } { w _ { i } } , \frac { \left[ \triangle y _ { b r } \right] _ { i j } } { h _ { i } } , \frac { w _ { j } h _ { j } } { w _ { i } h _ { i } } \right]$.
Then, we concatenate the above absolute and relative location feature as the spatial representations of the proposal, which is a 30-dim vector: $\widetilde { v } _ { l } ^ { i } = \left[ v _ { l }^{i} ; \delta v _ { l }^{i} \right] $. Finally, the subject features of each proposal is $ { v } _ { s } ^ { i } = [ { v } _ { s } ^ { i }; \widetilde { v } _ { l } ^ { i } ]$

\textbf{Object feature} is extracted for representing the object proposals with CNN and another spatial feature.
It is composed of C4 feature $v_{ij}=f_{CNN}(r_{j})$ and its relative location feature to subject. The relative location feature is encoded as follows: $\delta m _ { i j } = \left[ \frac { \left[ \triangle x _ { t l } \right] _ { i j } } { w _ { i } } , \frac { \left[ \triangle y _ { t l } \right] _ { i j } } { h _ { i } } , \frac { \left[ \triangle x _ { b r } \right] _ { i j } } { w _ { i } } , \frac { \left[ \triangle y _ { b r } \right] _ { i j } } { h _ { i } } , \frac { w _ { j } h _ { j } } { w _ { i } h _ { i } } \right]$. The object feature is $ { v } _ { o } ^ { ij } =  \left[ v _ { i j } ; \delta m _ { i j } \right]$. 

Noting that global visual features are not utilized in our method, as global features might introduce some ambiguity to the grounding task \cite{DBLP:conf/eccv/YuPYBB16}. 
\subsubsection{Referring Expression Features}
\label{reffeats}
The language features are extracted through LSTM \cite{DBLP:journals/neco/HochreiterS97}. 
Given an query $q = \left\{ w _ { t } \right\} _ { t = 1 } ^ { T }$,  first each word in $q$ is one-hot encoded and mapped into a word embedding $e_t$. Then the word embedding $e_t$ is fed into a bi-directional LSTM. The final representation $h_t = [\overrightarrow{h}_t, \overleftarrow{h}_t]$ is the concatenation of the hidden vectors in both directions. 

\subsection{Knowledge Extraction}
\label{knowledge}
We propose knowledge extraction module to introduce prior knowledge to fulfill the lack of annotations for subject and object.
The prior knowledge is in the form of semantic similarity between proposal category and the subject/object, which can guide the selection of subject and object proposals. The details are as follows.

For each proposal, we use the pre-trained Faster R-CNN to predict its category $C_i$. For the referring expression, we parse it into seven attributes: category name, color, size, absolute location, relative location, relative object, and generic attribute based on \cite{DBLP:conf/emnlp/KazemzadehOMB14}.
The category name is considered as the subject $W_s$, and the relative object is viewed as the object $W_o$ for each referring expression.

Next, we utilize an English vocabulary of 72,704 words contained in the GloVe pre-trained word vectors \cite{DBLP:conf/emnlp/PenningtonSM14} to encode the proposal category($C_i$) and subject($W_s$)/object($W_o$) into a vector. ``unk'' is used to indicate for the word which is out of the vocabulary. We calculate the cosine distance of the category and subject/object as semantic similarities.
\begin{equation}
\label{subscore}
\begin{aligned}
emb_s &= GloVe(W_s) \\
emb_o &= GloVe(W_o) \\
emb_c &= GloVe(C_i) \\                   
SIM_s  &= cos(emb_c, emb_s)\\
SIM_o  &= cos(emb_c, emb_o)
\end{aligned}
\end{equation}

\begin{figure}
	\centering
	\includegraphics[width=0.47\textwidth]{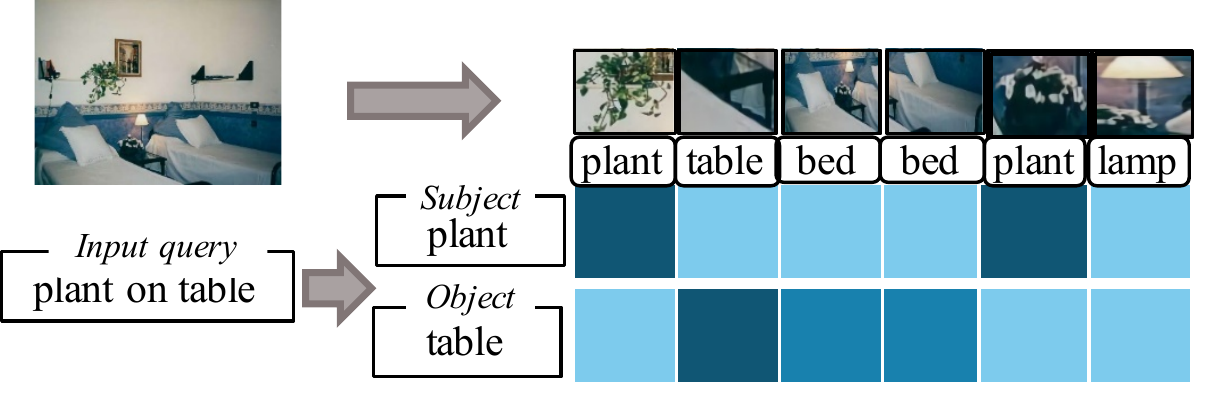}
	\caption{The knowledge extraction module.  Subject and object are extracted from the query. Categories are obtained for each proposal. We calculate the semantic similarity as the prior knowledge to guide the training of subject and object attention.}
	\label{similarity}
\end{figure}

The visualization of the semantic similarity is shown in Fig. \ref{similarity}

\subsection{Subject and Object Attention}
\label{sub_att}
Under the guidance of prior knowledge, we design subject and object attention to construct the subject-object proposal pairs.
Through subject attention, the candidate proposals with little probability being the target, are assigned to lower weights or even excluded in the next processor. Through object attention, the candidate proposal with the highest probability being the contextual entity is chosen as the object. 

This process is shown in Fig. \ref{structure} (b) and (c). $\widetilde { v } _ { s } ^ { i }$, $ v_{ij}$ are the subject and object features extracted from the proposals in the image. $emb_s$ and $emb_o$ represent the embedding of subject and object from the query. 
Taking the subject as an example, $\widetilde { v } _ { s } ^ { i }$ and $emb_s$ are first concatenated into one vector. Then the vector is fed into the subject attention, which is a two layer perceptron, to get the corresponding matching score between proposals and subject. The biases are omitted in Eq. (\ref{subscore}).
\begin{equation}
\begin{aligned}
\overline{Score}_s^{i}&= f_{ATT}\left(\widetilde { v } _ { s } ^ { i } ,emb_s\right) &=&\space W _ { 2 } \phi_{\rm ReLU} \left( W _ { 1 } [ \widetilde { v } _ { s } ^ { i } ,emb_s] \right)\\
\overline{Score}_o^{{ij}}&= f_{ATT}\left(v_{ij},emb_o\right) &=&\space W _ { 2 } \phi_{\rm ReLU} \left( W _ { 1 } [ v_{ij},emb_o] \right)
\end{aligned}
\end{equation}
We normalize the scores using softmax.
\begin{equation}
\begin{aligned}  
Score_s^i & = \operatorname { softmax } _ { i } \left( \overline {Score}_s^i\right)\\
Score_o^{ij} & = \operatorname { softmax } _ { i } \left( \overline{Score}_o^{ij} \right) 
\end{aligned}
\end{equation}

Furthermore, the semantic similarity of subject and object are considered as supervision for the two attention networks.
We use Mean Squared Error (MSE) criterion to minimize the distance between the matching score and the semantic similarity.
\begin{equation}
\begin{aligned}
Loss_{sub} &={\rm MSE}\left(Score_s, SIM_s\right)\\
Loss_{obj} &= {\rm MSE}\left(Score_o, SIM_o\right)
\end{aligned}
\end{equation}

Based on subject attention score, we have two different methods to obtain the candidate subject proposals. One is called as \textbf{soft filter}, which assigns different weights to the composed proposal pairs when grounding the query. The other is named as \textbf{hard filter}, which discards the unrelated subject proposals if their subject attention score is under the threshold.

Then, we choose the proposal with the maximum object attention score as the object proposal. This is due to the syntax habit that people tend to use the entity with a small number of occurrences in the image as a context to describe the target entity.

Finally, the proposal pairs are constituted based on the above subject proposals and object proposal. 

\subsection{Pairwise Attention}
\label{pair_att}
Different from previous methods, we try to find the best-matched pair $\{q, (r_s, r_o)\}$. $q$ denotes the query, $r_s$ denotes the subject proposal, and $r_o$ denotes the object proposal.  For each proposal pair $(r_s, r_o)$, a matching score between it and its corresponding query is learned through pairwise attention and an adaptive weighting scheme. The pair with the maximum matching score will be viewed as the final grounding result.

The procedure of pairwise attention is shown in Fig. \ref{structure}~(d). ${ v } _ { s } ^ { i }$ and $ { v } _ { o } ^ { ij}$ are the subject and object features of proposals. $h_t$ indicates the language feature extracted from the query through bi-directional LSTM. $ { v } _ { s } ^ { i }$, $ { v } _ { o } ^ { ij}$ and $h_t$are first concatenated into one vector. Then the vector is fed into the proposal pair attention to get the corresponding matching score between $(r_s, r_o)$ and $q$. The biases are omitted in Eq. (\ref{pairscore}).

\begin{equation}
\label{pairscore}
{ { Score } } _ { pair }^{i} = f _ { A T T } \left( h_t ,  { v } _ { s } ^ { i },  { v } _ { o } ^ { ij}\right) = W _ { 2 } \phi_{\rm ReLU} \left( W _ { 1 } [ h_t ,  { v } _ { s } ^ { i },  { v } _ { o } ^ { ij}] \right) 
\end{equation}

Further, because people are used to describing the target with the near contextual entity, we designed an adaptive weighting scheme for refining the correspondence based on the spatial relationship in the subject-object proposal pair. 
The weights are calculated as follows:
\begin{equation}
\omega_{dis}^{ij} = 100/(dist_{M}^{ij}+100),
\end{equation}
where $dist_M$ denotes the Manhattan distance between subject and object proposal in each proposal pair. 

The final score $S_t^i$ indicates the matching probability of proposal pair $(r_s, r_o)$ and query \textit{q}. 

If we use soft filter for the selection of subject proposals, the final score is 
\begin{equation}
\label{finalscore}
{ S } _ { t } ^ { i } = \operatorname {softmax } _ { i } \left( \omega_{dis}^{ij}\times Score_s^i \times { Score } _ { pair } ^ { i } \right),
\end{equation}
where $Score_s^i$ is the subject attention score learned according to prior knowledge.

If using hard filter, we directly discard the candidate pairs $(r_s, r_o)$, whose subject similarity is under the setted threshold. The matching score for the remaining proposal pairs is calculated as follows.
\begin{equation}
{ S } _ { t } ^ { i } = \operatorname {softmax } _ { i } \left( \omega_{dis}^{ij} \times  { Score } _ { pair } ^ { i } \right) 
\end{equation}

\subsection{Pairwise Reconstruction}
\label{rec}
We introduce a pairwise reconstruction to formulate the measurement of the grounding in the weakly supervised training stage, where the attentive proposal pairs are exploited to reconstruct the input query. 

First, the concatenated subject and object features of the proposal pair (${ r } _ { s } $, ${ r } _ { o } $ ) are fed into a one-layer perceptron.
\begin{equation}
r_{vis}^i=\phi _ {\rm R e L U } \left(W_v( \left[  { v } _ { s } ^ { i },  { v } _ { o } ^ { i j} \right])+b_v  \right)
\end{equation}

Then we  obtain the fused features  $f_{vis}$ according to the final matching score $S _ { t } ^ { i }$.
\begin{equation}
{ f } _ { vis } = \sum _ { i = 1 } ^ { N } S _ { t } ^ { i }  { r } _ { vis } ^ { i }
\end{equation}

Finally, inspired by the query generation methods \cite{DBLP:conf/cvpr/DonahueHGRVDS15, DBLP:conf/cvpr/VinyalsTBE15}, we reconstruct the input query through LSTM.
\begin{equation}
P ( q | f_{vis} ) = f _ {\rm L S T M } \left(f_{vis} \right)
\end{equation}
The language reconstruction network aims to maximize the likelihood of the ground-truth query $\hat { q }$ generated by LSTM,
\begin{equation}
\label{lanloss}
Loss _ { lan } = - \frac { 1 } { B } \sum _ { b = 1 } ^ { B } \log ( P ( \hat { q } | f_{alan} ) )
\end{equation}
where $B$ is the batch size.

\subsection{Attribute Classification Loss}
As mentioned in previous methods \cite{DBLP:conf/iccv/YaoPLQM17,DBLP:journals/pami/WuSWDH18, DBLP:conf/cvpr/Yu0SYLBB18}, attribute information is important on distinguish object of the same category.  Thus we also add an attribute classification branch in our model. The attribute label is extracted through an external language parser \cite{DBLP:conf/emnlp/KazemzadehOMB14} according to \cite{DBLP:conf/cvpr/Yu0SYLBB18}.
Subject feature $\widetilde { r } _ { s } ^ { i }$ of proposal is used for attribute classification.
As each query has multiple attribute labels, we use the binary cross-entropy loss for the multi-label classification.

\begin{equation}
Loss _ { a t t  } = f_{\rm BCE} \left( y _ { i j }, p_{ij} \right)
\end{equation}
We use the reciprocal of the frequency that attribute labels appears as weights in this loss to ease unbalanced data.

\subsection{Network Training and Inference}
The network is trained with an end-to-end strategy. During training, only query with attribute words goes through attribute classification branch. At inference, the reconstruction module is not needed anymore. We feed the image and query into the network and get the most related proposal pair whose final score is the maximal in the grounding module.
\begin{equation}
j = \arg \max _ { s } f  \left( p , (r _ { s }, r_o) \right)
\end{equation}

The final loss function is:
\begin{equation}
Loss = Loss_{sub} + Loss_{obj} + Loss_{lan} +Loss_{att}
\end{equation}

\section{Experiments}

\begin{table*}[]
	\centering
	\caption{Accuracy (IoU $>$ 0.5) on RefCOCO, RefCOCO+ and RefCOCOg. \textbf{Bond}: best result.  \bluefont{Blue}: best result of VC. }

	\scalebox{1.3}{
		\begin{tabular}{c|c|ccc|ccc|c}
			\hline
			\multirow{2}{*}{Methods}&\multirow{2}{*}{Settings}&\multicolumn{3}{c|}{RefCOCO}&\multicolumn{3}{c|}{RefCOCO+}& RefCOCOg\\ 
			\cline{3-9}&&val&testA&testB&val&testA&testB& val\\ \hline 
			VC  & w/o reg& - &13.59& 21.65& - & 18.79 & 24.14 & 25.14 \\ 
			VC & -& - & 17.34 & 20.98 & - & 23.24 & 24.91 & \bluefont{33.79} \\ 
			VC  & w/o $\alpha$ & - & \bluefont{33.29} & \bluefont{30.13} & - & \bluefont{34.60} & \bluefont{31.58} & 30.26 \\ \hline
			VC  (det) & w/o reg & - & 17.14 & 22.30 & - & 19.74 & 24.05 & 28.14 \\ 
			VC (det) &- & - & 20.91 & 21.77 & - & 25.79 & 25.54 & 33.66 \\ \
			VC (det) & w/o $\alpha$  & - & 32.68 & 27.22 & - & 34.68 & 28.10 & 29.65 \\ \hline \hline
			
			KPRN & soft & 34.43&33.82
			&35.45&35.96&35.24
			&36.96& 33.56\\
			
			KPRN &  soft + attr&\textbf{36.34}&\textbf{35.28}&\textbf{37.72}&\textbf{37.16}&\textbf{36.06}&\textbf{39.29}&36.65\\
			
			KPRN & hard &35.04 &34.74&36.53&35.10&32.75&36.76&35.44\\ 
			
			KPRN & hard +attr&34.93&33.76&36.98&35.31&33.46&{37.27}&\textbf{38.37}\\ \hline
			
	\end{tabular}}
	\label{refcoco}
\end{table*}
\subsection{Datasets}
We evaluate our method on four popular benchmarks of referring expression grounding.

\noindent \textbf{RefCOCO \cite{DBLP:conf/eccv/YuPYBB16}.} 
It is also called UNC RefExp.
The dataset contains 142,209 queries for 50,000 objects in 19,994 images from MSCOCO \cite{DBLP:conf/eccv/LinMBHPRDZ14}.
It is collected through ReferitGame \cite{DBLP:conf/emnlp/KazemzadehOMB14}.
The dataset is split into train, validation, Test A, and Test B, which has 16,994, 1,500, 750 and 750 images, respectively. Test A contains multiple people while Test B contains multiple objects. Each image contains at least 2 objects of the same object category. 
The average length of the queries in this dataset is 3.61.

\noindent \textbf{RefCOCO+ \cite{DBLP:conf/eccv/YuPYBB16}.} 
It has 141,564 queries for 49,856 referents in 19,992 images from MSCOCO \cite{DBLP:conf/eccv/LinMBHPRDZ14}. 
It is also collected through ReferitGame \cite{DBLP:conf/emnlp/KazemzadehOMB14}. 
Different from RefCOCO, the queries in this dataset are disallowed to use locations to describe the referents. Thus, this dataset focus on the appearance of the referents.
The split is 16,992, 1,500, 750 and 750 images for train, validation, Test A, and Test B respectively. Each image contains 2 or more objects of the same object category in this dataset. The average length of the queries in this dataset is 3.53.

\noindent \textbf{RefCOCOg \cite{DBLP:conf/cvpr/MaoHTCY016}.} It is also called Google Refexp. 
It has 95,010 queries for 49,822 objects in 25,799 images from MSCOCO \cite{DBLP:conf/eccv/LinMBHPRDZ14}. This dataset is collected in a non-interactive setting on Amazon Mechanical Turk. Thus 
It has longer queries containing appearance and location to describe the referents. 
The split is 21,149 and 4,650 images for training and validation. There is no open test split for RefCOCOg. Images were selected to contain between 2 and 4 objects of the same category. The average length of the queries is 8.43.

\noindent \textbf{RefCLEF \cite{DBLP:conf/emnlp/KazemzadehOMB14}.} It is also called ReferIt. 
It contains 20,000 annotated images from IAPR TC-12 dataset \cite{article} and SAIAPR-12 dataset \cite{DBLP:journals/cviu/EscalanteHGLMMSPG10}. The dataset includes some ambiguous queries, such as anywhere. It also has some mistakenly annotated image regions. 
The dataset is split into 9,000, 1,000 and 10,000 images for training, validation and test for fair comparison with \cite{DBLP:conf/eccv/RohrbachRHDS16}. 
100 bounding box proposals \cite{DBLP:conf/cvpr/HuXRFSD16} are provided for each image using Edge Boxes \cite{DBLP:conf/eccv/ZitnickD14}. Images contain between 2 and 4 objects of the same object category. The maximum length of all the queries is 19 words.

%

\subsection{Experimental Setup}
\subsubsection{Implementation details}
We train KPRN through Adam \cite{DBLP:journals/corr/KingmaB14} with an initial learning rate 4e-4, which drops by 10 after every 8,000 iterations. The training iterations are 30,000 with a batch size of a single image. Each image has an indefinite number of annotated queries. The rectified linear unit (ReLU) \cite{DBLP:conf/nips/MontufarPCB14} is used as the non-linear activation function. Batch normalization operations are not used in our framework. 
ResNet is the main feature extractor for RoI visual features. 
We adopt EdgeBoxe \cite{DBLP:conf/eccv/ZitnickD14} to generate 100 region proposals for RefCLEF dataset for fair comparison with \cite{DBLP:conf/eccv/RohrbachRHDS16, DBLP:conf/cvpr/ChenGN18}. 

\subsubsection{Metrics}
The Intersection over Union (IoU) between the selected region and the ground-truth are calculated to evaluate the network performance. If the IoU score is greater than 0.5, the predicted region is considered as the right grounding.

\subsection{Results on RefCOCO, RefCOCO+ and RefCOCOg datasets}
\subsubsection{Performance Analysis:}
We compared the proposed KPRN with the only published unsupervised results on these datasets \cite{DBLP:conf/cvpr/ZhangNC18}. 
``soft'' denotes that we use soft filter for the selection of subject proposals. ''hard'' denotes that we discard the candidate subject proposals whose attention score is under the threshold. ``attr'' denotes that whether to use attribute classification loss in the model.

Table \ref{refcoco} reports the  comparison results on RefCOCO, RefCOCO+ and RefCOCOg datasets. 
We can have the following findings. First, our method performs better on testB, which can outperform VC by a large margin on RefCOCO and RefCOCO+ datasets. While on testA, the promotion is less. The difference between testA and testB is that testA contains multiple people and testB contains multiple other objects. 
Second, the KPRN with soft filter achieves better results on RefCOCO and RefCOCO+ datasets, while KPRN with hard filter performs better on RefCOCOg dataset. 
Third, the attribute classification loss can improve the performance by about 2\% on average on these datasets.

\subsubsection{Ablation Study:}
\begin{table*}[]
	\centering
	\caption{Ablation study on RefCOCO, RefCOCO+ and RefCOCOg datasets. }
	\scalebox{1.2}{
		\begin{tabular}{l|ccc|ccc|c}
			\hline
			\multirow{2}{*}{Methods}&\multicolumn{3}{c|}{RefCOCO}&\multicolumn{3}{c|}{RefCOCO+}& RefCOCOg\\ 
			\cline{2-8}&val&testA&testB&val&testA&testB& val\\ \hline             
			KPRN (attr)  &15.58&8.34&25.57&16.25&14.15&18.41&33.64\\
			KPRN (attr+loc)  &36.47&35.11&37.74&20.33&17.34&24.38&36.13\\ 
			KPRN (attr+loc+obj) &36.73&35.60&37.11&36.49&35.92&38.41&37.17\\ 
			KPRN (attr+loc+obj+hard) &35.31&33.87&36.17&34.78&33.34&37.43&40.24\\
			KPRN (attr+loc+obj+soft) &35.28&35.41&36.60&35.36&35.16&36.29&38.45\\
			KPRN (attr+loc+obj+hard+dist) &35.98&35.28&37.39&35.66&33.67&38.52&43.16\\
			KPRN (attr+loc+obj+soft+dist) &{36.34}&{35.28}&{37.72}&{37.16}&{36.06}&{39.29}&36.65\\\hline
			
	\end{tabular}}
	\label{refcoco_ablation}
\end{table*}
We study the benefits of each module by running ablation experiments.
Table \ref{refcoco_ablation} reports the results on RefCOCO, RefCOCO+ and RefCOCOg datasets with different settings. ``attr'', ``loc'', ``obj'', ``soft'', ``hard'' and  ``dist'' denotes the visual attribute features, location features, context features, soft filter for subject attention, hard filter for subject attention and weighting scheme for spatial distance, respectively.  

Some observation can be obtained as follows.
First, location features play a critical role on RefCOCO dataset. On RefCOCO+ dataset, the performance does not increase significantly with location features, that is probably because this dataset is disallowed to use locations to describe the referents.
Second, the performance increases by a large margin on RefCOCO+ dataset when introducing the object features of proposal. This indicates the effectiveness of proposed object attention.
Third, subject attention can increases the performance of RefCOCOg dataset. 
Fourth,  the adaptive weighting scheme for spatial distance is also helpful for the grounding results.

Table \ref{refcoco_thr} represents results of hard filter for the selection of subject proposals with different filter threshold. The performance increases with the increase of the threshold on RefCOCOg, while the threshold has weak impact on RefCOCO and RefCOCO+ datasets.
\subsubsection{Qualitative Results}
\begin{figure*}[tp]
	\centering
	
	\includegraphics[width=0.95\textwidth]{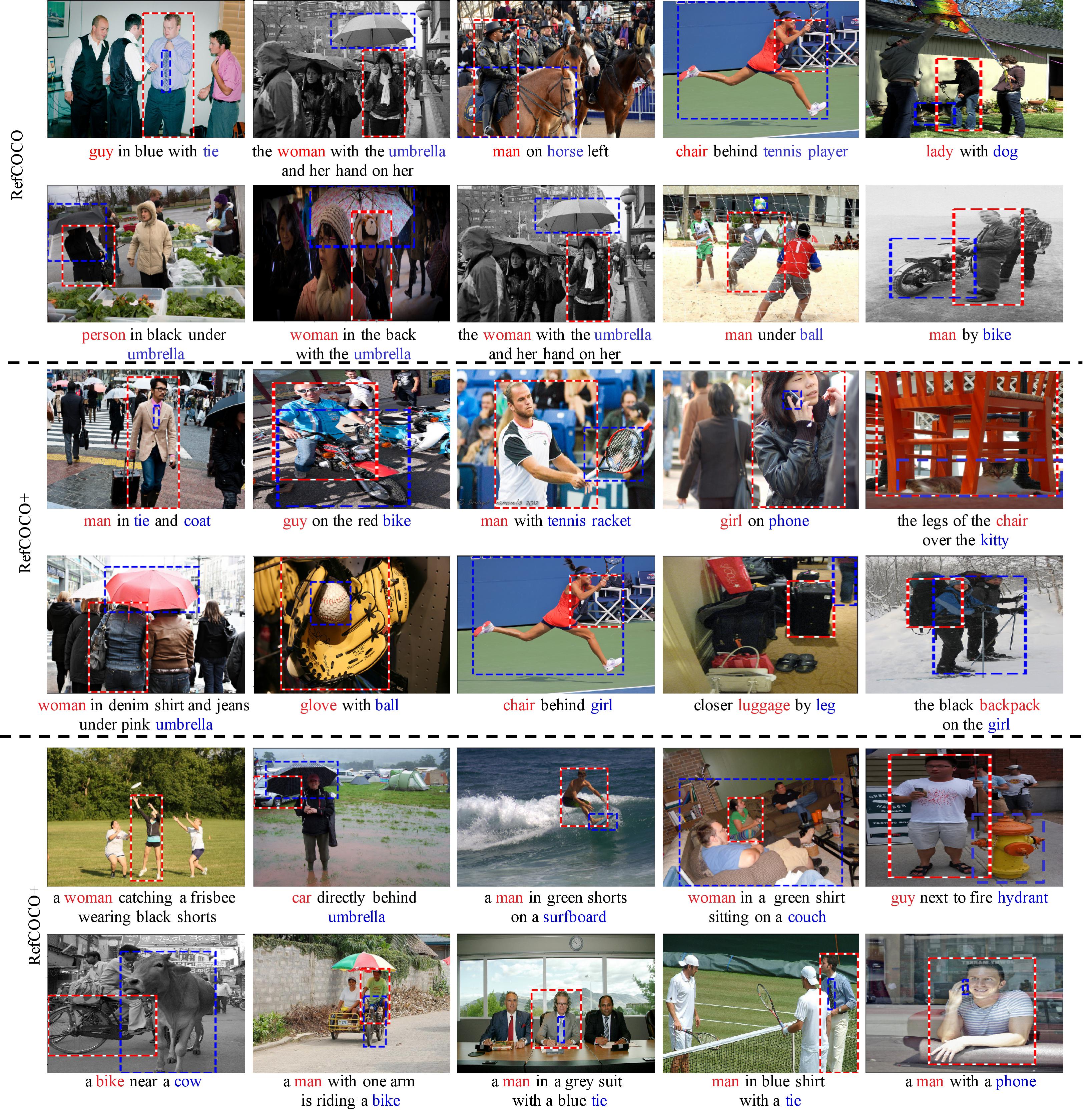}
	\caption{Qualitative results on RefCOCO, RefCOCO+ and RefCOCOg datasets. The denotations of the bounding box colors are as follows. Solid white: ground truth; dashed red: predicted subject proposal; dashed blue: predicted object proposal. }
	\label{visualization}
	
\end{figure*}

Fig. \ref{visualization} shows qualitative example predictions on RefCOCO, RefCOCO+ and RefCOCOg datasets. The query is shown under the corresponding image. The ground truth, grounding proposal and context proposal are denoted as solid white, dashed red and dashed blue, respectively.

\subsection{Results on RefCLEF Dataset}
\begin{table}[]
	\centering
	\caption{Albation study on different hard filter threshold for RefCOCO, RefCOCO+ and RefCOCOg datasets.}
	\scalebox{0.9}{
		\begin{tabular}{c|c|ccc|ccc|c}
			\hline
			\multirow{2}{*}{expID}&\multirow{2}{*}{thr}&\multicolumn{3}{c|}{RefCOCO}&\multicolumn{3}{c|}{RefCOCO+}& RefCOCOg\\ 
			\cline{3-9}&&val&testA&testB&val&testA&testB&val\\ \hline 
			
			exp1  &  0.05 &35.92&35.28&36.76&36.60&34.65&38.35&36.74\\        
			exp2 &   0.10   &34.93&33.76&36.98&35.31&33.46&37.27&38.37 \\
			exp3 & 0.15&33.96&32.31&36.78&32.56
			&28.94&36.51&37.35\\
			exp4& 0.20&34.68&33.07&37.90&32.60&29.60&36.53&39.26\\exp5& 0.25&34.89&33.83&37.45&32.66&30.53&36.74&41.13\\exp6&0.30&35.53&35.23&37.10&33.71&32.19&36.74&42.03\\
			exp7&0.35&35.98&35.28&37.39&35.66&33.67&38.52&43.16\\
			exp8&0.40&35.24&34.17&37.84&35.84&34.13&39.58&40.93\\

			\hline
			
	\end{tabular}}
	
	\label{refcoco_thr}
\end{table}

\begin{table}[]
	\centering
	\caption{Accuracy (IoU $>$ 0.5) on RefCLEF dataset.}
	\scalebox{1.1}{    \begin{tabular}{l|l}
			\hline
			Method  & IoU  \\ \hline
			LRCN \cite{DBLP:conf/cvpr/DonahueHGRVDS15}   & 8.59     \\ 
			Caffe-7K \cite{DBLP:conf/rss/GuadarramaRSZFD14} & 10.38    \\ 
			GroundeR \cite{DBLP:conf/eccv/RohrbachRHDS16}& 10.70    \\ 
			MATN  \cite{DBLP:conf/cvpr/0006LZF18} & 13.61    \\ 
			VC \cite{DBLP:conf/cvpr/ZhangNC18} & 14.11\\
			VC w/o $\alpha$ \cite{DBLP:conf/cvpr/ZhangNC18} & 14.50\\
			KAC Net \cite{DBLP:conf/cvpr/ChenGN18} & 15.83    \\ \hline \hline
			KPRN (attr+loc+obj) &  20.99\\
			KPRN (attr+loc+obj+soft) &  18.35\\
			KPRN (attr+loc+obj+hard) &  32.32\\
			KPRN (attr+loc+obj+hard+dist)& \textbf{33.87} \\ \hline
	\end{tabular}}
	\label{refclef}
\end{table}
\subsubsection{Performance Analysis:}
We compare our KPRN with state-of-the-art weakly supervised referring expression grounding methods. 
The LRCN use the image captioning to score how likely the query phrase is to be generated for the proposal box. CAFFE-7K predicts a class for each proposal box and then compared to the query phrase after both are projected to a joint vector space. Other methods are all introduced in the related work.

Table \ref{refclef} reports the results on RefCLEF dataset. We can have the following observations. 
First, the proposed KPRN outperforms state-of-the-art result with a large margin of 18.04\%. 
Second, by introducing extra location features and object proposal features, our method improves the IoU by 5.16\%.
Third, when using soft filter for subject proposals, the performance degrades for grounding. When using the hard filter, the performance obtains an improvement of 12.88\%. This is probably because there are too many candidate proposals in this dataset. With the power of subject attention, many unrelated proposals can be excluded. Thus the number of candidate proposal pairs for grounding can be decreased by a large margin.
Fourth,  Through adding the adaptive weighting scheme, the performance of KPRN gets a further promotion of 1.55\%, which benefits from the modeling of spatial relationship in the subject-object proposal pair.

\subsubsection{Ablation Study:}
We also verify the impact of the threshold for hard filter. The results in Table \ref{refclef_ablation} demonstrate that the IoU gets the best result when the threshold is 0.5.
\begin{table}[]
	\centering
	\caption{Ablation study on different hard filter threshold for RefCLEF dataset.}
	\scalebox{1}{\begin{tabular}{c|ccccccc}
			\hline
			expID  & exp1&exp2&exp3&exp4&exp5&exp6 \\ \hline
			thr & 0.1&0.2&0.3&0.4&0.5&0.6\\ 
			IoU & 19.48 &23.32 &33.17&33.87&35.98&35.78\\ 
			\hline
			
	\end{tabular}}
	
	\label{refclef_ablation}
\end{table}

\section{Conclusion}
To address the weakly supervised REG, we propose a knowledge-guided pairwise reconstruction network (KPRN). 
The KPRN can model the relationship between subject and object as well as ground them under the guidance of prior knowledge.
The prior knowledge is obtained in the form of semantic similarities between each proposal and the subject/object.
Then, we design the subject and object attention to construct the subject-object proposal pairs with the supervision of such knowledge.
Further, Pairwise attention and a weighting scheme are used to learn the final matching score between proposal pairs and query.
Finally, pairwise reconstruction measures the grounding performance under weakly supervised setting.
Experiments demonstrate that the proposed method provides a significant improvement of performance on four datasets.

\begin{acks}
This work was supported in part by National Natural Science Foundation of China: 61771457, 61732007, 61772494, 61672497, 61622211, 61836002, 61472389, 61620106009 and U1636214, in part by Key Research Program of Frontier Sciences, CAS: QYZDJ-SSW-SYS013, and Fundamental Research Funds for the Central Universities under Grant WK2100100030.
\end{acks}

%
\bibliographystyle{ACM-Reference-Format}
\bibliography{MM-sigconf}

\end{document}